# A deep convolutional neural network model for rapid prediction of fluvial flood inundation

Syed Kabir[1,2]*, Sandhya Patidar[2], Xilin Xia[1], Qiuhua Liang[1], Jeffrey Neal[3] and Gareth Pender[2]

[1]School of Architecture, Building and Civil Engineering, Loughborough University, Loughborough, United Kingdom.

x.xia2@lboro.ac.uk ; q.liang@lboro.ac.uk

[2]School of Energy, Geoscience, Infrastructure and Society, Heriot-Watt University, Edinburgh, United Kingdom.

s.patidar@hw.ac.uk ; g.pender@hw.ac.uk

[3]School of Geographical Sciences, University of Bristol, Bristol, United Kingdom

j.neal@bristol.ac.uk

* Corresponding author

Syed Kabir

School of Architecture, Building and Civil Engineering

Loughborough University

Loughborough, Leicestershire, LE11 3TU

Email: S.R.Kabir@lboro.ac.uk

## Abstract

Most of the two-dimensional (2D) hydraulic/hydrodynamic models are still computationally too demanding for real-time applications. In this paper, an innovative modelling approach based on a deep convolutional neural network (CNN) method is presented for rapid prediction of fluvial flood inundation. The CNN model is trained using outputs from a 2D hydraulic model (i.e. LISFLOOD-FP) to predict water depths. The pre-trained model is then applied to simulate the January 2005 and December 2015 floods in Carlisle, UK. The CNN predictions are compared favourably with the outputs produced by LISFLOOD-FP. The performance of the CNN model is further confirmed by benchmarking against a support vector regression (SVR) method. The results show that the CNN model outperforms SVR by a large margin. The CNN model is highly accurate in capturing flooded cells as indicated by several quantitative assessment matrices. The estimated error for reproducing maximum flood depth is 0 ~ 0.2 meters for the 2005 event and 0 ~ 0.5 meters for the 2015 event at over 99% of the cells covering the computational domain. The proposed CNN method offers great potential for real-time flood modelling/forecasting considering its simplicity, superior performance and computational efficiency.

**Keywords**

Rapid flood modelling; deep learning; convolutional neural network; machine learning; flood inundation

## 1. Introduction

Two-dimensional (2D) hydraulic/hydrodynamic models have been widely applied to simulate complex hydrological processes and flood dynamics. Recent advancements in computing technology along with the increasing availability of high-resolution remotely sensed data, such as terrain elevation and river morphology, have enabled these sophisticated models to be applied at the regional to global scales (e.g., Yamazaki *et al.*, 2011, de Paiva *et al.*, 2013). However, due to their high computational demand, it is still challenging to use these physically-based sophisticated models for operational real-time flood forecasting (Bhola *et al.*, 2018).

Considerable research effort has been devoted to improving the overall performance of hydraulic/hydrodynamic models for large-scale flood modelling. For example, the computational efficiency of these models may be improved by implementing parallel computing algorithms to take advantages of multiple processors (Neal *et al.*, 2018, Sanders & Schubert, 2019). Xia *et al.* (2019) developed a new fully hydrodynamic modelling framework that utilizes the state-of-the-art high-performance graphics processing units (GPUs) for modelling fluvial flooding from rainfall-generated overland flow to inundation at a high spatial resolution across a large catchment of 2500 $km^2$. However, despite the advances in high-performance computing technology and the development of subsequent computational methods, significant challenges still exist regarding the application of 2D hydrodynamic models for operational flood forecasting. One of the key challenges is that, while it is now possible to make a single model run in real-time using a GPU-accelerated hydrodynamic model (e.g., Ming *et al.*, 2020), it is still computationally prohibited to run such a model multiple times using ensemble numerical weather predictions to provide reliable flood forecasts with an acceptable lead time.

An alternative approach is to adopt an offline method for operational flood inundation mapping as proposed in Bhola *et al.* (2018). The proposed offline method requires the construction of a database using pre-run inundation maps and river discharges. During or before a flood event, the inundation maps with a matching discharge are requested and retrieved from the database to provide forecasts (Leedal *et al.*, 2010, Bhola *et al.*, 2018). Although such a system does not require live 2D simulations, preparing the database of inundation maps can be labour intensive and requires storage of a large volume of data. Furthermore, due to changing environment, e.g., land-use change, geomorphological change and engineering construction, the flood scenarios may become outdated and new simulations are required to regularly update the database, creating extra effort and resources for maintenance.

The ideal solution to these various technical challenges would be to develop a new modelling approach that can relax the mentioned computational burden while still being able to generate practically useful and statistically significant information to support real-time flood forecasting. A possible solution would be to use machine learning (ML) models that can closely emulate the outputs of 2D hydrodynamic models. Application of ML techniques for rainfall-runoff forecasting has been investigated for a few decades. In contrast, research on the application of ML for flood inundation modelling remains very limited and only a handful of such ML approaches have been reported to date. For example, in Chang *et al.* (2010, 2014), Jhong *et al.* (2018) and Shen & Chang (2013), hybrid ML techniques were successfully used for flood modelling. Chang *et al.* (2018a) developed a self-organizing map (SOM), an artificial neural network (ANN) designed for clustering operations, integrated with recurrent nonlinear autoregressive exogenous (R-NARX) networks for flood modelling at regional scale, giving flood forecasting of up to 12h ahead in the Kemaman River Basin, Malaysia. This approach was further improved in Chang *et al.* (2018b) by developing an intelligent hydroinformatics integration platform (IHIP) to provide a user-friendly web interface for enhanced capability in online flood forecasting and risk management. Of particular relevance to the current study, Liu & Pender (2015) developed an inundation model based on a support vector regression (SVR) algorithm for emulating the outputs of a fine grid model (FGM). In their approach, SVRs were first trained using a small number of the outputs from the FGM and then applied to predict water depths and velocities at the target locations. Bermúdez *et al.* (2019) presented a least squared-support vector machine (LS-SVM) method to compute the spatial distribution of the maximum water depth and velocity in a coastal urban area using three sets of flow and tidal data. Berkhahn *et al.* (2019) proposed an ensemble neural network method to predict in real-time the maximum water levels of a flash flood event induced by spatially uniform rainfall. A data-driven 3h ahead fluvial flood inundation mapping framework was proposed by Kabir *et al.* (2020), which shows that the wet/dry condition of the cells inside the simulation domain is a function of the upstream discharge magnitude and duration. However, this study only focused on classifying wet/dry cells to indicate flooded areas in a small English town of ~ 5 $km^2$ and the flooding processes was only driven by one river channel. These studies have demonstrated that the computationally much less expensive ML approaches may be used to efficiently predict flood variables once they are trained and calibrated appropriately.

When selecting an ML technique for flood modelling, we should bear in mind that most of the existing algorithms are not suitable for multi-output scenarios, i.e. predicting a flood variable (e.g., depth) in multiple cells through a single model. In this regard, artificial neural networks (ANNs) may potentially be more useful as it can be implemented to solve both of the single-output and multi-output problems. Over the last decade, convolution neural networks (CNNs) have gained unprecedented success in solving computer vision problems due to their ability to extract unknown features and learn compact representations and represent the current leading deep learning (DL) paradigm. Nevertheless, the potentiality of CNNs in modelling high-resolution flood inundation has not yet been tested and confirmed.

In this work, we present an innovative rapid fluvial flood modelling approach based on CNNs to predict spatially distributed water depths for two large flood events in the City of Carlisle, UK. This is for the first time that the capability of a one-dimensional CNN (1D-CNN) model, accelerated by modern GPUs, is investigated and tested for rapid prediction of fluvial flooding in a domain containing over half of a million cells. The rest of the paper is organised as follows: Section 2 introduces methodology including the proposed CNN modelling framework; the experimental setup and application are explained in Section 3; Section 4 presents and discusses the results; and finally, brief conclusions are drawn in Section 5.

## 2. Methodology

This section explains how the CNN-based fluvial flood modelling framework is developed for reproducing two historical flood events in the City of Carlisle, UK.

### *2.1 Overall research strategy*

Fig. 1 illustrates the five steps in developing and assessing the new CNN model for fluvial flood inundation modelling:

Step 1: Generate sufficient number of synthetic hydrographs to represent different hydrological and flood conditions at the upstream boundary points of the three river channels feeding to the case study site;

Step 2: Use these hydrographs as the boundary conditions to drive a 2D-hydraulic/hydrodynamic model (e.g. LISFLOOD-FP as adopted in this study) to create spatially distributed inundation sequences and subsequently derive the time-series of water depth at each cell in the domain for the entire flood duration;

Step 3: Generate training and validation datasets for the candidate DL model (e.g. CNN model in this work); for training, the upstream synthetic hydrographs are used as input variables and the hydraulic simulation outputs (i.e. water depth sequences) are used as the target variable;

Step 4: Develop, train and optimize the candidate DL model;

Step 5: Use observed or predicted upstream hydrographs to drive the trained DL model to predict fluvial flood inundation (e.g. to reproduce the Carlisle 2005 and 2015 floods in this work) and assess its performance.

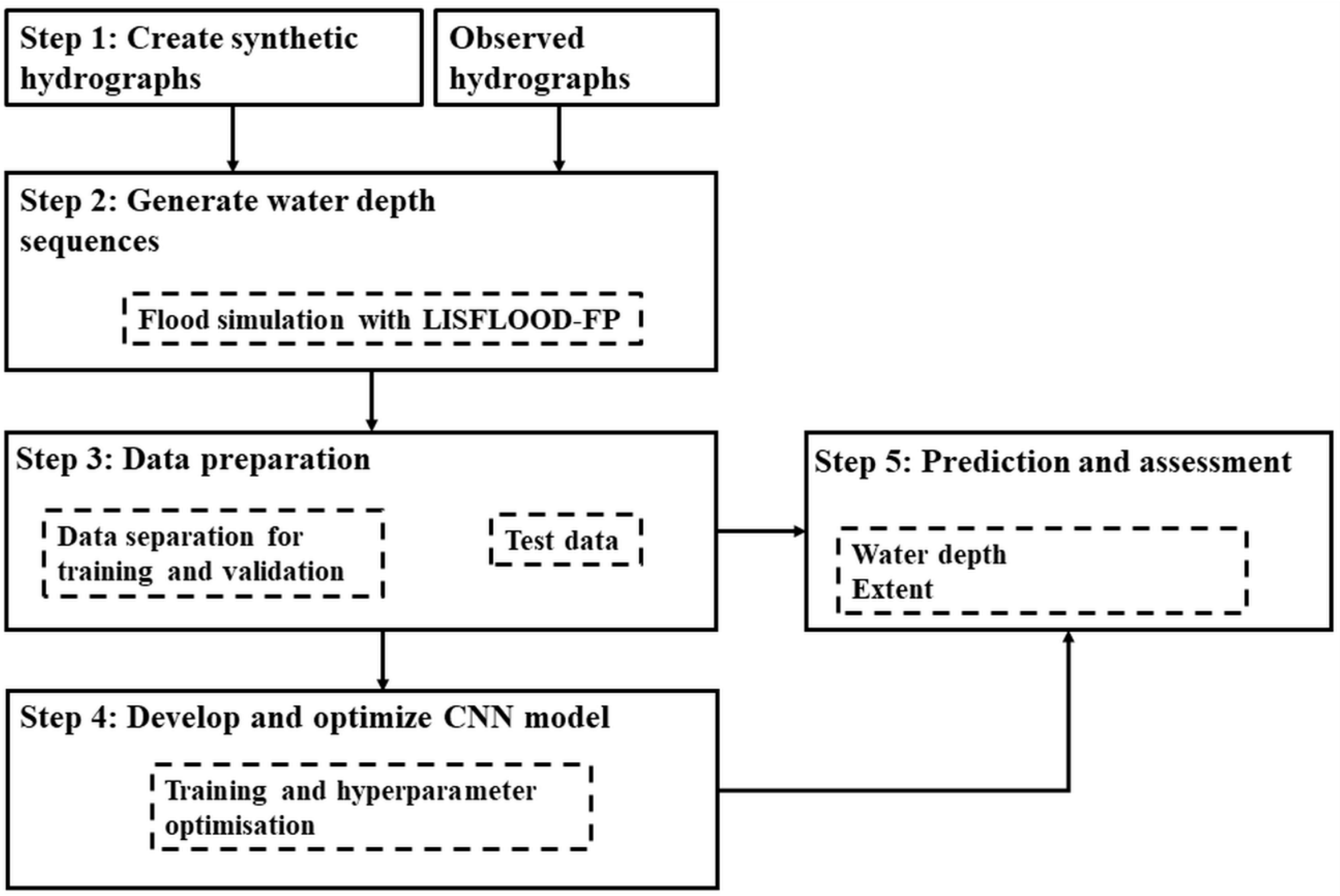

Figure 1. The core steps in developing and testing the proposed CNN flood model.

To assess the capability of the proposed CNN model in emulating the results of a 2D hydraulic model, the DL predictions in terms of water depths are directly compared with the outputs from the adopted hydraulic model, i.e. LISFLOOD-FP. In addition, the performance of the CNN model will be further evaluated by comparing with an SVR approach, one of the popular ML methods used in earlier studies (e.g., Lin *et al.*, 2013, Liu & Pender, 2015, Chang *et al.*, 2018c, Bermúdez *et al.*, 2019). A point-wise comparison is made between the models at 18 pre-selected control points (13 of which are taken from the list of locations where water depths were surveyed by the EA and University of Bristol after the 2005 event) that represent the main flood risk zone. Several assessment matrices are adopted here to quantify the prediction errors.

The root mean squared error (*RMSE*) (Barnston, 1992) is defined as

$$RMSE = \sqrt{\frac{\sum_{i=1}^{N}(O_i - P_i)^2}{N}} \qquad (1)$$

where $N$ is the sample size, $O_i$ and $P_i$ are the 'observed' and 'predicted' values, respectively. *RMSE* = 0 returns a perfect fit between the predicted and 'observed' data.

The Nash-Sutcliffe efficiency (*NSE*) coefficient (Nash & Sutcliffe, 1970) may be calculated as follows

$$NSE = 1 - \frac{\sum_{i=1}^{N}(O_i - P_i)^2}{\sum_{i=1}^{N}(O_i - \bar{o})^2} \qquad (2)$$

where $\bar{o}$ is the mean of 'observed' data. The value of *NSE* generally varies between 0 and 1, where *NSE* = 1 represents a perfect fit between the reference and predicted data and a negative *NSE* indicates that the model fails to reproduce the test case.

To evaluate the accuracy in predicting flood inundation, the *precision*, *recall* and *F1* metrics (Eq. 3-5) are also considered to demonstrate how precisely the models predict 'true positives' (cells correctly predicted as flooded by the predictive models) to the total predicted positives (the sum of the cells correctly and wrongly predicted as flooded) and to the total actual positives (cells predicted as flooded by the LISFLOOD-FP) (Bermúdez *et al.*, 2019) over the entire domain:

$$Precision = \frac{True\ positive}{Total\ predicted\ positive} \tag{3}$$

$$Recall = \frac{True\ positive}{Total\ actual\ positive} \tag{4}$$

$$F1 = 2 \times \frac{Precision \times Recall}{Precision + Recall} \tag{5}$$

A higher *precision* value implies that most of the cells predicted as flooded by the predictive models are also classified as flooded by the reference model, i.e. LISFLOOD-FP. A higher *recall* value means that cells classified as flooded are well captured by the predictive models. *F1* is the harmonic mean of *recall* and *precision* metrics and defines how well the model predictions match the reference results, with a score of 1 returning a perfect match.

It is worth noting that the primary objective of this study is to investigate the capability of the CNN model in emulating the outputs of a 2D hydrodynamic/hydraulic model. Therefore, the outputs of the LISFLOOD-FP model rather than the field observations are used as the reference to assess the predictive performances of the ML models under consideration.

### *2.2 Overview of models being used*

In this section we describe the hydraulic and the ML models adopted to develop, train and assess the proposed rapid flood modelling system.

#### *2.2.1 Convolutional Neural Networks (CNNs)*

Since its inception in 1990, CNNs have become a research hotspot and the *de facto*

standard for various ML and computer vision (CV) applications. In particular, the annual ImageNet large scale visual recognition challenge (ILSVRC) 2012 has changed the course of image classification problems through the application of deep CNNs. However, the key factor which has made the deep CNNs extremely popular is the ever-increasing computational power of the modern computers.

CNNs are, in general, feed-forward neural networks with alternating convolutional and subsampling layers and are predominantly trained in a supervised manner (Kiranyaz *et al.*, 2019). Deep CNNs have been exclusively developed to operate on 2D data (images and videos) and commonly known as '2D-CNNs'. 2D-CNNs can extract features and learn complex objects from large volume of labelled data. Whilst the classical CNNs were developed specifically for 2D signals, Kiranyaz *et al.* (2015) proposed the first 1D-CNN to handle sequential data (1D signal). Since then, 1D-CNNs have gained significant popularity and been successfully applied in various fields, e.g., biomedical data classification (Zihlmann *et al.*, 2017), structural damage detection (Abdeljaber *et al.*, 2018) and sentiment analysis (Munandar *et al.*, 2018).

### *2.2.2 LISFLOOD-FP flood inundation model*

The physically-based LISFLOOD-FP hydraulic model is used to generate training samples for the data-driven predictive models considered in this work. First reported by Bates & De Roo (2000), LISFLOOD-FP is a raster-based model for simulating fluvial or coastal flood inundation. The model has been improved significantly over the last two decades and tested successfully in numerous case studies across the globe (e.g., Knijff *et al.*, 2010, Amarnath *et al.*, 2015, Komi *et al.*, 2017).

The early version of LISFLOOD-FP solves the zero-inertial approximation of the Saint Venant equations (i.e. diffusion-wave approximation) using an explicit forward difference scheme on a staggered grid over a 2D plane (Bates and De Roo, 2000). Specific to the present study, the LISFLOOD-FP version 6.3.1 is used, which was implemented with an ‘acceleration’ solver to recover the inertial terms from the Saint Venant equations (Bates *et al.*, 2010, de Almeida *et al.*, 2012) for calculating floodplain inundation and a ‘sub-grid’ solver (Neal *et al.*, 2012) for representing channelized flows below the adopted grid resolution. The technical details of the model can be found in Bates *et al.* (2010). The model and relevant documentation can be acquired from http://www.bristol.ac.uk/geography/research/hydrology/models/lisflood/.

### *2.2.3 Support vector regression (SVR)*

The SVR is a kernel-based supervised learning method developed from the original support vector machine (SVM) method (Cortes & Vapnik, 1995) to solve regression problems. The basic feature of SVR is to map the input space into a high-dimensional feature space using a non-linear mapping function (for solving non-linear problems), through which the non-linearity of input vectors becomes linearly separable (Raghavendra and Deka, 2014).

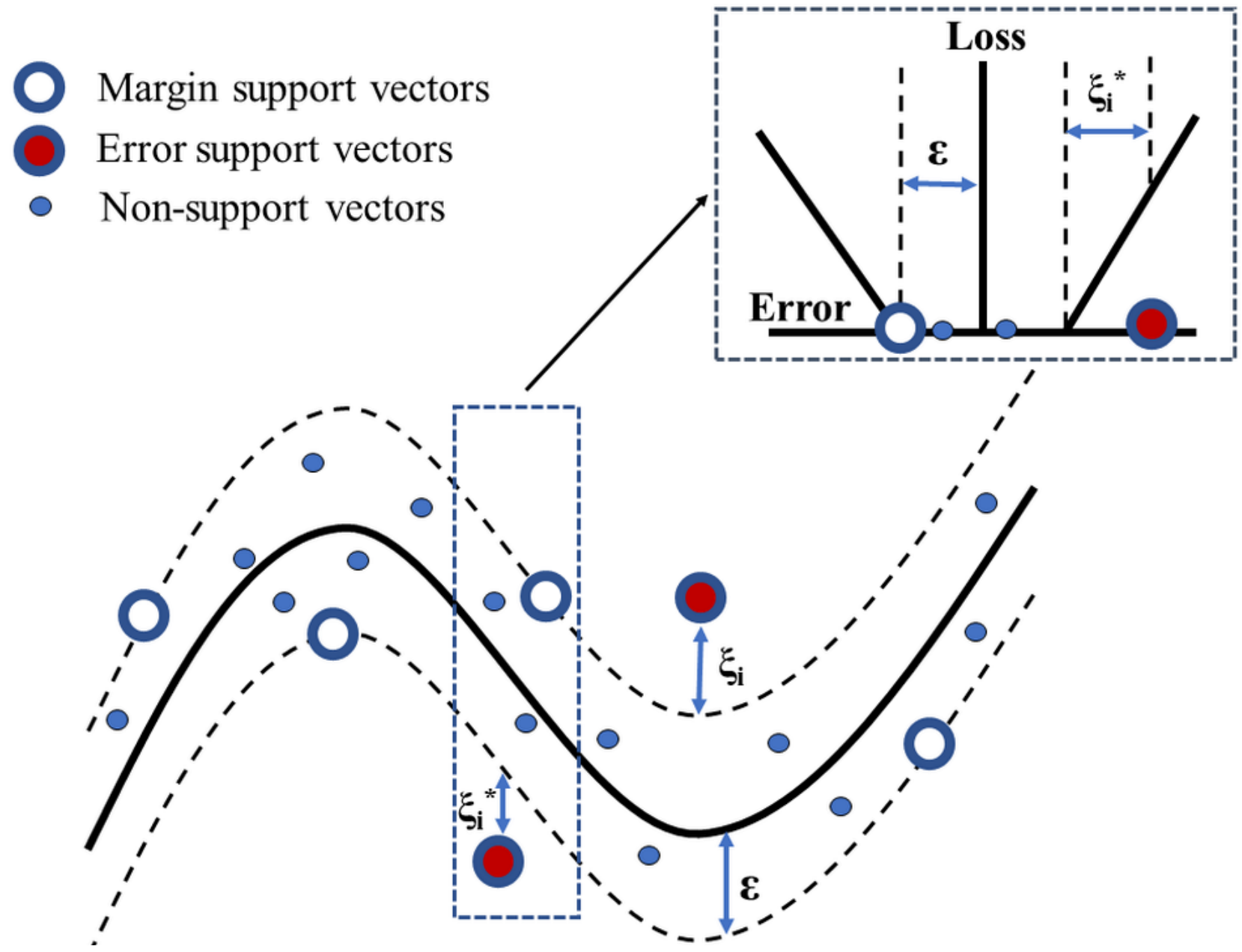

Figure 2. Non-linear SVR with an $\varepsilon$-insensitive zone in the feature space.

The concept of non-linear SVR in the feature space is illustrated in Figure 2, in which $\xi_i$ and $\xi_i^*$ are the so-called slack variables and these non-negative variables are introduced to estimate the deviation of training data samples lying outside of the $\varepsilon$-insensitive zone. The effectiveness of an SVR model depends upon several mutually linked parameters, e.g., kernel parameters, cost parameter and the width of the $\varepsilon$-insensitive zone. The cost parameter is a positive constant indicating the smoothness of the approximation function. A large cost parameter may overfit the training data but an inappropriately small value may lead to underfitting. Parameter $\varepsilon$ controls the width of the insensitive zone, influencing the number of support vectors and thus the overall generalization ability of an SVR model.

In this study, to draw a fair comparison between the CNN and SVR flood models, identical training and testing datasets are used. The same parameter optimisation technique is also employed to search for appropriate model hyperparameters.

## 3. Model setup and application

This section gives the key details related to how LISFLOOD-FP, CNN and SVM models are setup and applied, and the data being used.

### *3.1 Case study*

To demonstrate the performance of the CNN flood model, the city of Carlisle is used as the case study in this work and the site-specific information and key hydrometric and spatial data are introduced herein.

### *3.1.1 Site description*

Located in the downstream Eden Catchment in the Northwest England, the city of Carlisle is highly prone to flooding. The total drainage area of Eden Catchment is nearly 2500 km$^2$ and is fed by an average precipitation of approximately 1148 mm/yr (Allen *et al.*, 2010). The study domain covers about 14.5 km$^2$ of the urbanised area of Carlisle (Fig. 3), which has been heavily hit by historical flood events including the 2005 and 2015 floods. There are three main river channels within the area of interest, i.e. River Eden, Petteril and Caldew, which drive the flood dynamics. The locations (within the study domain) at high risk of flooding are at the confluence of the Rivers Eden and Caldew at Willow Holme, and the confluences of Little Caldew, River Petteril and River Eden at Durranhill, Botcherby and Warwick Road.

In 2005, the River Eden and its two tributaries (River Caldew and River Petteril) caused unprecedented flooding on the 8$^{th}$ January due to persistent rainfall that started on the 7$^{th}$ January (Roberts *et al.*, 2009). The onset of the event was slow with initial flooding occurring in the early hours of the day, which was well before the peak that arrived around noon. The inundation predominantly occurred in the residential/commercial zones along the channels and low-lying rural areas situated on the north-eastern part of the city. The annual exceedance probability (AEP) of the event was estimated to be 0.59%, corresponding to a 170-year return period (Cumbria County Council, 2017). Following this flood event which affected approximately 1600 properties, flood defences were constructed to protect the high-risk areas of the city for flood events similar to or greater than the one in 2005 (AEP 0.5%). However, on 5$^{th}$ and 6$^{th}$ December 2015, the Storm Desmond brought in record-breaking rainfall that subsequently led to widespread flooding across the Eden Catchment, with Carlisle again being heavily hit. The magnitude of the event (AEP 0.33%) exceeded the designed criteria of the Carlisle flood defence scheme (Cumbria County Council, 2017). The flooding followed by overtopping of the embankments and flood walls devastated the city and its people, affecting approximately 2100 properties. The flow rate recorded at the Sheepmount gauging station in River Eden in the morning of the 6th December was 1680 $m^3$/s, breaking all the past records. These two major events provide an opportunity to test whether a model performs consistently under different conditions, i.e. with and without flood defences. For this reason, both of the 2005 and 2015 flood events are used to test the proposed CNN model.

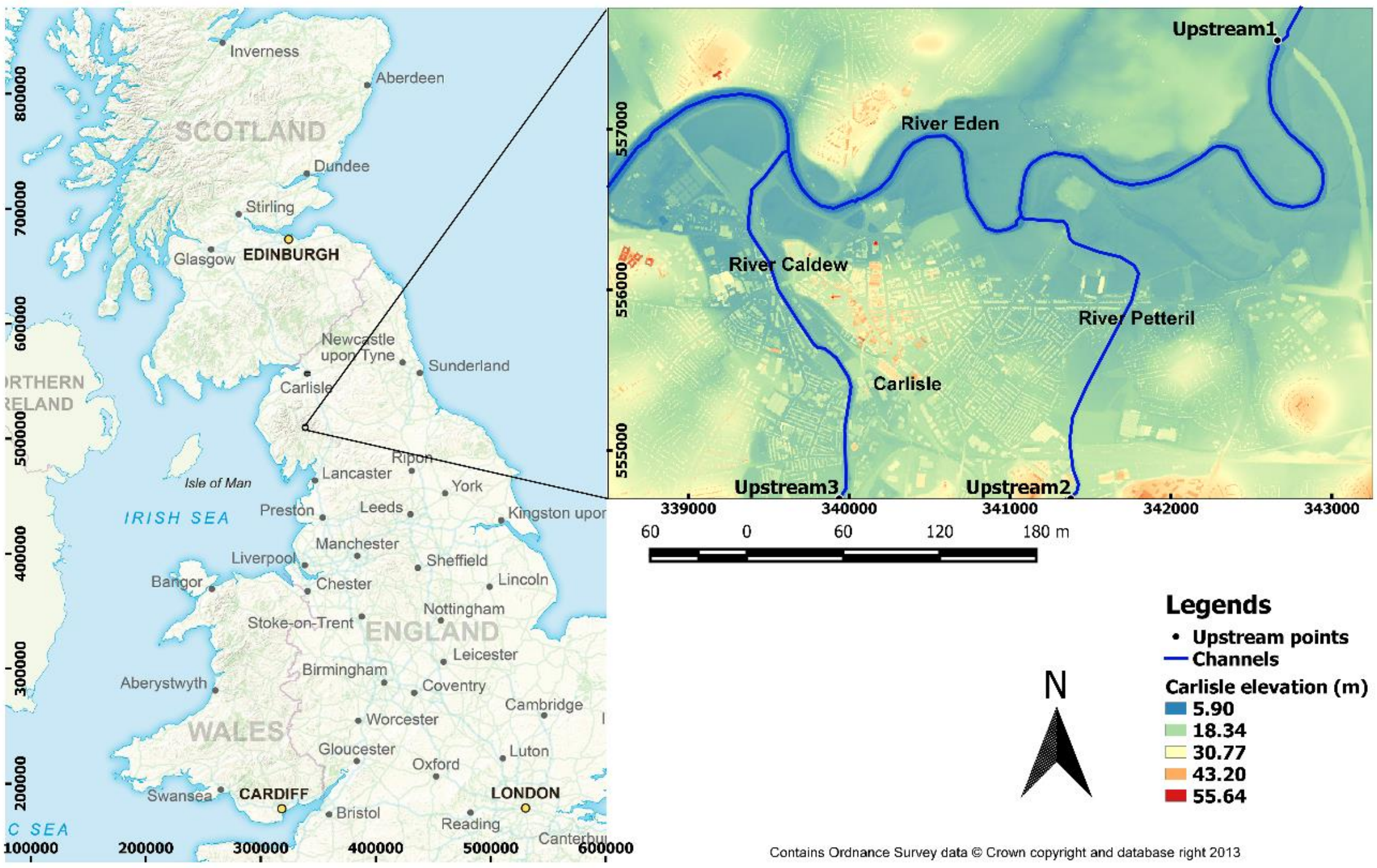

Figure 3. Study domain and topography.

### *3.1.2 Hydrometric and topographic data*

Hydrometric and topographic data are required to set up and run LISFLOOD-FP for predicting flood inundation. Essential data to reproduce the Carlisle flood events include upstream boundary conditions (e.g. flow hydrographs) and digital elevation model (DEM) for the case study site. In this work, we use the DEM provided by the School of Geographical Sciences at the University of Bristol, which was created by further processing the LiDAR DEM provided by the UK Environment Agency (EA) by removing the noises and obstacles (e.g. trees and bridges). The DEM has a 5m spatial resolution and was deemed sufficient to reproduce the flood dynamics using LISFLOOD-FP in the previous studies (e.g. Neal *et al.*, 2013). However, this DEM is only valid for the 2005 flood event because new flood defences have been built afterwards. To model the 2015 event and investigate the impact of flood alleviation scheme, the data in the format of a shapefile are acquired from the government open data platform (https://data.gov.uk/) for the defences built between 2005 and 2015 and then embedded in the DEM for reproducing the 2015 event, in which all of the flood defences are assigned with a constant height of 2m.

To drive the two flood events, boundary conditions are needed at the upstream points of the three feeding rivers. The three boundary points are marked in Fig. 3 as "Upstream 1" (Easting: 342682, Northing: 557532) located under the M6 bridge in the North-East of the domain, and "Upstream 2" (Easting: 341362, Northing: 554702) and "Upstream 3" (Easting: 339947, Northing: 554702) located at the south edge of the domain. Herein, the coordinate reference system OSGB 1936 (EPSG:27700) is adopted. It is worth noting that these upstream boundary points do not coincide with flow observation gauges in the rivers and therefore no direct observation data are available to drive the flood simulations. For the 2005 event, 15-minute discharge hydrographs obtained through 1D flood routing are available from previous studies (e.g. Neal *et al.*, 2013, Parkes *et al.*, 2013) and are used in this study without further modification. For the 2015 event, the 15-minute measured hydrographs from the nearby gauging stations, i.e. Great Corby in Eden River, Harraby Green Business Park in Petteril and Cummersdale in Caldew are directly imposed in the respective boundary points to drive the flood simulations.

### *3.2 Generating input hydrographs*

To reproduce the 2005 and 2015 Carlisle floods and demonstrate the performance of the ML-based predictive models, it is necessary to generate sufficient input (discharge) and output (water depth) data to train the ML models. For this purpose, 24 'synthetic' hydrographs (8 for each of the three upstream boundary locations) with various peaks and durations are produced to represent flood scenarios of different magnitudes. In each of the upstream locations, hydrographs of the historical floods with smaller peaks than the January 2005 event are selected and adjusted using the following formula to increase their magnitudes as appropriate:

$$Q_n = Q_{obs} \times \frac{Peak_{max}}{Q_{max}} \qquad \{Peak_{max} > Q_{max}\} \qquad (6)$$

In which $Q_n$ is the synthetic flow discharge, $Peak_{max}$ is the user specified peak, $Q_{obs}$ is the observed discharge and $Q_{max}$ is the observed peak discharge. This generates the synthetic hydrographs A-E as shown in Figure 4. In addition, three more hydrographs (hydrographs F-H in Figure 4) are selected from the historical records at different gauging stations of River Eden to complete the training datasets (for all three upstream points). When generating or selecting the hydrographs, the flow in the River Eden is ensured to be substantially larger than the flows in the two tributaries to reflect the reality. The final 8 x 3 'synthetic' hydrographs are presented in Figure 4, together with the routed/measured hydrographs to drive the 2005 and 2015 flood simulations. A brief summary of these hydrographs is provided in Table 1.

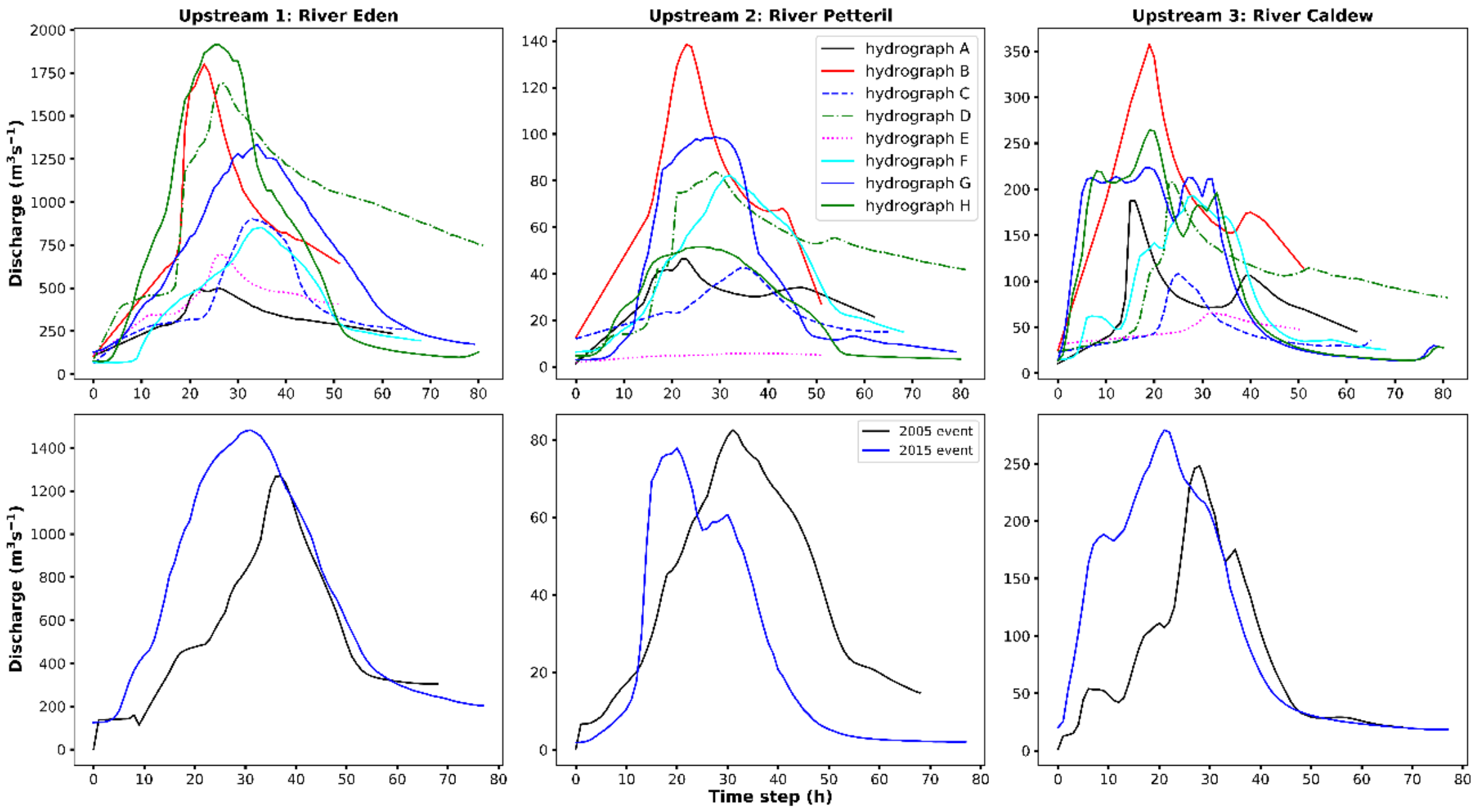

Figure 4. The 'synthetic' and routed/observed hydrographs used to train and test the ML models. For the 2005 Carlisle flood event, the hydrographs start at 00:00 hours on 7th January 2005 as time 0. The hydrographs for the 2015 event start at 23.15 hours on 4th December as time 0.

Table 1. Summary of the hydrographs used in this study.

| Name | Type | Purpose | Upstream 1 peak [$m^3/s$] | Upstream 2 peak [$m^3/s$] | Upstream 3 peak [$m^3/s$] |
|---|---|---|---|---|---|
| Hydrograph A | Synthetic | Training | 500.00 | 46.81 | 194.73 |
| Hydrograph B | Synthetic | Training | 1800.00 | 138.88 | 365.65 |
| Hydrograph C | Synthetic | Training | 900.00 | 42.59 | 109.20 |
| Hydrograph D | Synthetic | Training | 1700.00 | 83.68 | 209.63 |
| Hydrograph E | Synthetic | Training | 700.00 | 5.74 | 65.43 |
| Hydrograph F | Synthetic | Training | 854.28 | 82.57 | 193.33 |
| Hydrograph G | Synthetic | Training | 1338.07 | 99.16 | 226.74 |
| Hydrograph H | Synthetic | Training | 1950.50 | 51.99 | 266.91 |
| Event 2005 | Approximated | Testing | 1272.94 | 82.57 | 248.74 |
| Event 2015 | Approximated | Testing | 1486.19 | 78.90 | 279.30 |

### *3.3 Generating target data for model training and testing*

To generate training samples, LISFLOOD-FP is run sixteen times using the eight sets of synthetic inflow hydrographs to produce different flood conditions in the study site described by DEMs with and without embedding the flood defences constructed after 2005. The output files contain grid-based (raster format) water depths distributing over the study site at 15-minute temporal resolution. The test samples are generated respectively using the 2005 routed hydrographs on the pre-flood defence DEM and the 2015 approximate hydrographs on the post-flood defence DEM. For all of the simulations, the friction coefficient (Manning's $n$) is set to be uniform across the whole domain. Three different values (0.035 $sm^{-1/3}$, 0.045 $sm^{-1/3}$ and 0.055 $sm^{-1/3}$) are tested, and 0.055 $sm^{-1/3}$ produces reasonable results and hence is adopted to support the simulations to generate the training and test datasets.

Herein, a depth threshold of 0.3m is applied to negate the insignificant depths from the target data, i.e.

$$Y = \begin{cases} x & x > 0.3 \\ 0 & Otherwise \end{cases} \tag{7}$$

where $Y$ represents the water depth value in a stacked array (further description can be found in Section 3.6). This value is selected based on the flood hazard thresholds for different objects (e.g. 0.15m for transport links and 0.3m for buildings) as suggested by Aldridge *et al.* (2016) when building their impact library for flood risk assessment in the UK.

### *3.4 Constructing the predictive models*

In this work, the proposed CNN model is developed in Python programming language using Keras module within the Tensorflow 2.1 framework. More specifically, the sequential application programming interface (Keras Sequential API) is used to build the model, layer-by-layer. The core structure of the 1D-CNN model is graphically illustrated in Fig. 5. The network has five hidden layers, including two convolutional layers and three dense layers. The dense layers are fully connected and act like a multi-layer perceptron (MLP) network. The output layer contains nodes equal to the number of cells in the simulation domain (i.e. 581,061 for the current case study), and the input layer receives the upstream flow discharge values.

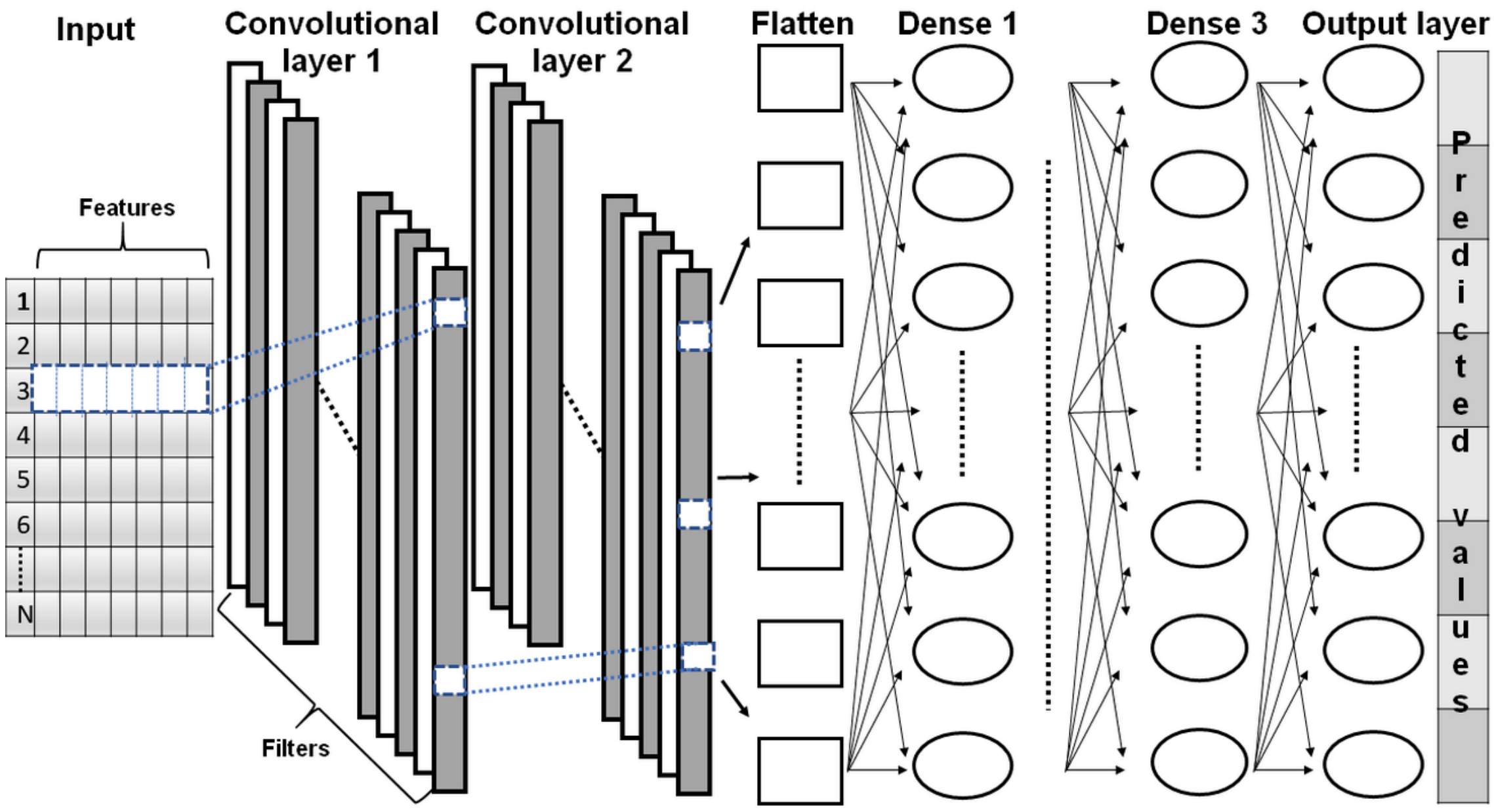

Figure 5. Structure of the adopted CNN for water depth prediction. Here, the 'features' contain the model inputs (i.e. time-series of discharge with lags and observation time), with each of the rows (1-N) containing information at one 'time-step'. The output of the model is an array of water depths which are then converted to spatially distributed flood maps.

In addition to the CNN, an SVR flood modelling method is also considered and constructed herein for comparison. Unlike CNN, SVR does not allow training and making predictions for all of the cells at one time and it is also not practical to calibrate an SVR model at each of the cells. Bermúdez *et al.* (2019) trained 25,000 individual SVR models and then linearly interpolated the predicted results to produce the distributed water depth maps across the whole simulation domain represented by a DEM of 1m spatial resolution. The area of the current study site is similar to that of Bermúdez *et al.* (2019) but the spatial resolution of the adopted DEM is 5m. Therefore, 500 locations are randomly selected and spatially distributed to represent the study domain, with each of these locations fitted by a separate SVR model. The regression kriging (RK) interpolation method (Hengl *et al.*, 2007) is then used to deduce the depths at the unsampled locations. The SVR models are built using the Scikit-learn library in Python. More details on parameter selection and optimisation are provided in the following sub-sections.

### *3.5 Hyperparameter optimisation*

The objective of hyperparameter optimisation is to find the 'best' set of parameters of a given model that lead to the optimum performance as measured on the validation data. In this study, the Bayesian optimisation approach is adopted to tune each of the predictive models. For either of the 2005 or 2015 events (simulated with different DEMs), we use the eight synthetic flood scenarios for hyperparameter optimisation and keep the 'real-event' data for model testing. We also optimise 18 SVR models for each of the control points, leading to 18 sets of SVR parameters. The optimised SVR models are then iteratively run using these parameter sets to find the global parameters by identifying the parameter set that produces the lowest error for all 18 control points. The use of a global parameter set is essential because optimising 500 SVR models is practically unfeasible. The final searched hyperparameter sets of the CNN and SVR models for the current case study are summarised in Table 2.

Table 2. Hyperparameters of the CNN and SVR predictive models.

| Model | Filter size | Neurons | Optimiser | Batch size |
|---|---|---|---|---|
| CNN | [32, 128] | [32, 256, 512] | Adam | 10 |
| | **Kernel** | **Cost** | **Epsilon** | **Gamma** |
| SVR | Radial basis | 25.296 | 0.031 | 0.016 |

### *3.6 Training the ML models and making predictions*

On each of the two DEMs for the study site, the CNN and the benchmark SVR models are trained using input (predictor) and output (target) variables from the eight synthetic flood scenarios (hydrographs A-H in Fig. 4). The training process is completed in two steps.

In the first step, the input and target variables are defined by converting the eight synthetic hydrographs at each of the upstream points and the corresponding raster-based flood depths predicted by LISFLOOD-FP into matrices. In this study, we explicitly use discharge values with eight antecedent time-steps (corresponds to 2h of LISFLOOD-FP model initialisation time) for each of the upstream locations and the corresponding observation time as the primary inputs for predicting water depths. This leads to a total of 28 input variables. The discharge with antecedent values and corresponding observation times for all eight synthetic hydrographs are stacked vertically to create the input feature matrix containing 2104 samples (Figure 6A). The target matrix is created by converting the sequential water depth raster files into arrays and then vertically stacking them together (Figure 6B). This results in a matrix of size 2104 X 581061, where 581061 is the total number of cells in the domain.

In the second step, the ML models are trained using the input and target variables. Each row in the input and output matrices is treated as a sample of the training dataset. During the CNN training process, different regularisation techniques, including ‘early stopping’ (stops the training process when the model performance does not improve after a certain number of iterations), ‘batch normalisation’ (Ioffe & Szegedy, 2015) and ‘dropout’ (Srivastava *et al.*, 2014) are applied to prevent the model from overfitting. It is found that, along with ‘early stopping’ criteria, ‘batch normalisation’ and ‘dropout’ (with dropout rate = 0.2) are particularly effective for reproducing the 2015 event for which the flood defences are considered in the DEM. Therefore, these regularisation techniques are used for the 2015 event. However, inclusion of ‘batch normalisation’ and ‘dropout’ did not increase the accuracy for 2005 event, and therefore, only ‘early stopping’ criteria is applied. The training process is repeated twice using the 2005 and 2015 training datasets (referred to as CNN-2005 and CNN-2015 models hereafter), respectively. Figure 6C sketches the training process for CNN-2015. The CNN hyperparameters presented in Table 2 is slightly adjusted for the CNN-2015 by increasing the batch size to 32. This is because, a general practice is to use a larger ‘batch size’ when ‘batch normalisation’ regularisation technique is applied.

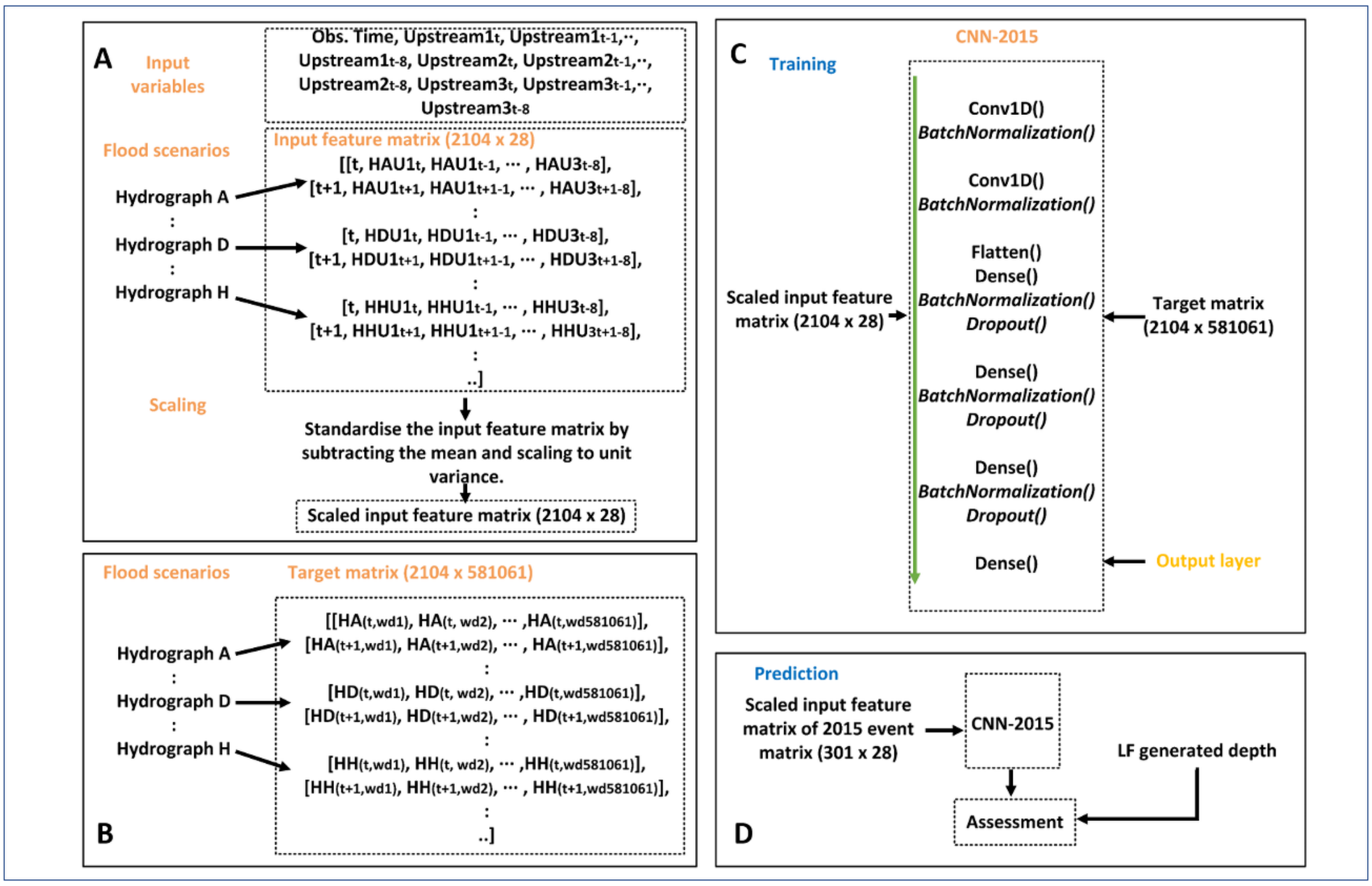

Figure 6. The proposed CNN inundation modelling framework. Here, for example, *t* is the flow ‘observation’ time, $HAU1_t$ and $HA_{(t,\ wd1)}$ are the corresponding flow rate at Upstream1 and water depth at cell number 1 for the Hydrograph A and the associated flooding scenario. The 2015 flood event has 301 discharge values (301 samples) and the 2005 event has 266 samples. The predicted outputs are compared with the LISFLOOD-FP (LF) generated results.

At this stage, the SVR models are also trained using the same set of input and target data at the sampled locations in sequence. That is to replace the CNN-2015 model with an SVR in Figure 6C and train a separate model in sequence for each of the sampled locations (500 in total). The global model parameters given in Table 2 are used during the SVR training process.

The CNNs are trained on an NVIDIA Tesla P100 GPU while the SVR models are trained using an Intel I5-9400 CPU with six cores. Finally, as illustrated in Fig. 6D, the test datasets (2005 and 2015 flood events) are ingested to the pre-trained CNN-2005, CNN-2015 and SVR models to predict water depths over the entire case study domain at a 15-minute time interval. The point-wise predictions from the SVR models are spatially interpolated using the RK method, as mentioned in Sect. 3.4. The results and performance of the models are evaluated using the assessment matrices introduced previously.

## 4. Results and discussion

To demonstrate the performance of the proposed CNN method, the 2005 and 2015 Carlisle floods are reproduced, and the results are compared with those produced by the LISFLOOD-FP and the SVR approach in this section.

### *4.1 Point-wise comparison between the CNN-2005, SVR and LISFLOOD-FP predictions for the 2005 event*

For the 2005 event, the three routed hydrographs as presented in Fig. 4 are imposed at the three upstream boundary points of the corresponding rivers to drive the three flood models under consideration. The CNN-2005 and SVR predictions in terms of time-series of flood depth are first compared with those predicted by the LISFLOOD-FP at 18 control locations as indicated in Fig. 7. The results are plotted in Fig. 8, in which a threshold of 0.3m is used to screen the depth values, i.e. the predicted depth is truncated to be 0 when it is smaller than 0.3m. Both ML models perform reasonably well in all the control points, but slightly more evident discrepancies can be noticed at two locations, i.e. Building 1 and Water mark 3. The reason may be due to the relatively shallow depth and short flood duration in these two locations. In other words, the flood states at these two locations change rapidly during the flood event, thus providing insufficient training samples for the ML models to learn this highly dynamic ‘feature’.

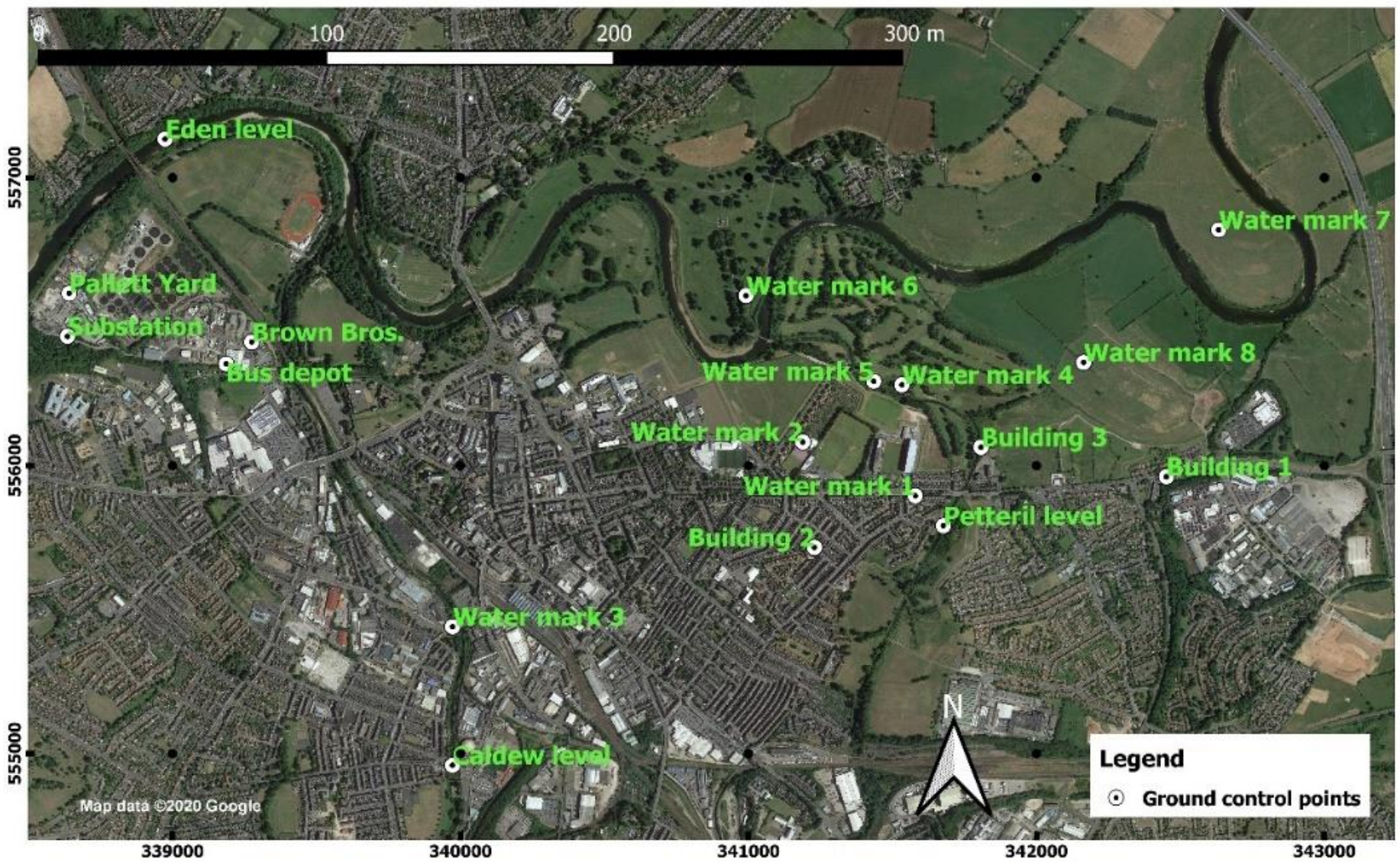

Figure 7. Control locations used for assessing model performance.

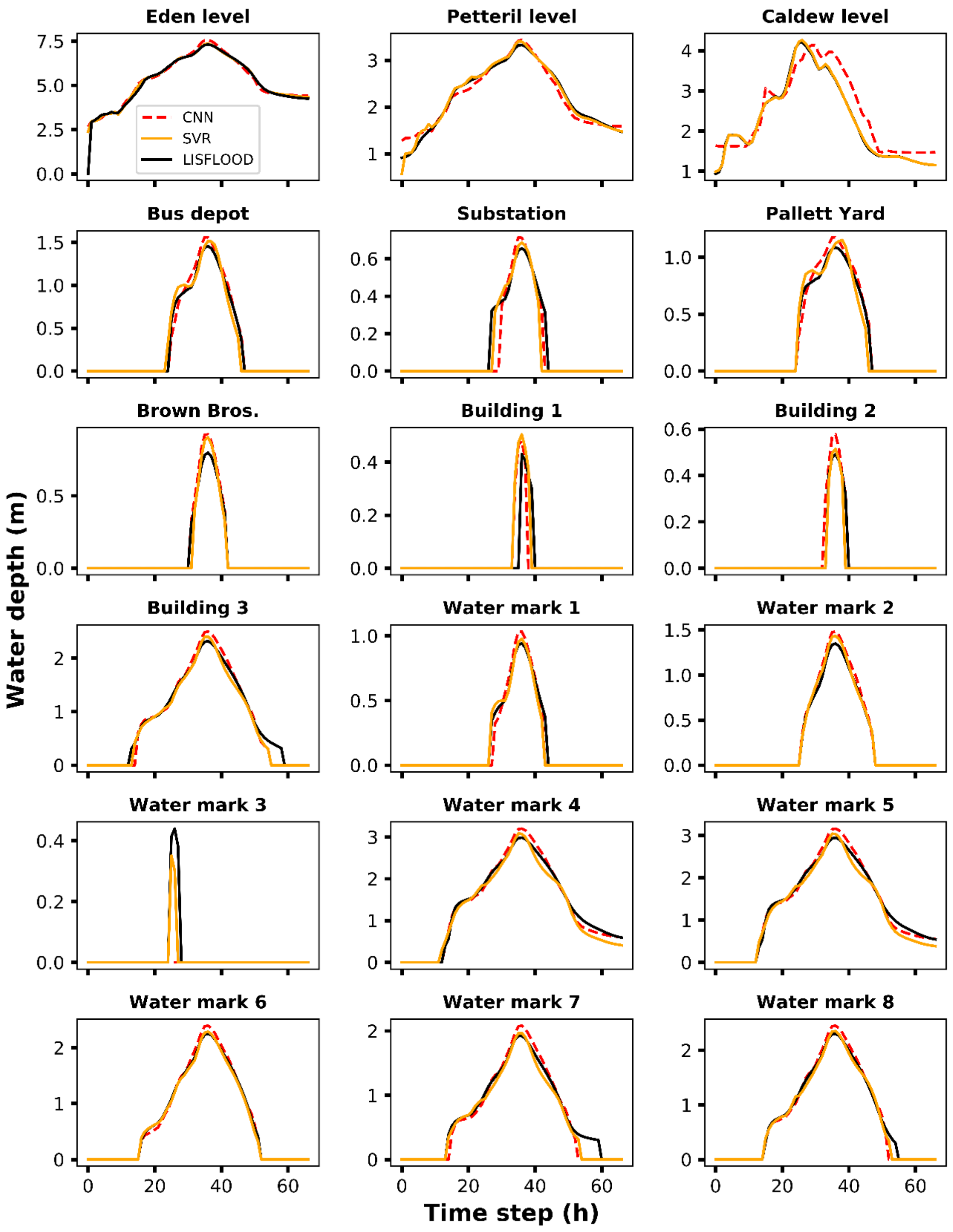

Figure 8. Comparing the time-series of water depths predicted by CNN-2005 and the SVR model against LISFLOOD-FP.

To quantitatively compare the ML predictions with the reference results from LISFLOOD-FP, the aforementioned error measures including NSE and RMSE are calculated at the 18 control locations and the mean values are listed in Table 3. Both of the ML models perform reasonably well in this case, i.e. returning high mean values for NSC and low mean values for RMSE. From these statistic matrices, the SVR model seems to perform slightly better than CNN-2005. For example, the CNN model returns a slightly lower mean NSE (0.86 vs. 0.91) than the SVR model. After removing the two control locations (Building 1 and Water mark 3) where the CNN model performs less satisfactorily from the statistics, the performance of the two models is shown to be both much improved and becomes similar (0.96 vs. 0.97 for CNN-2005 and SVR). Arguably, these locations are exposed to small water depth and short flood duration and therefore less important for overall flood risk management. As a whole, the satisfactory point-wise comparison shows that the ML models can aptly predict maximum depths, arrival, and receding times. However, it is worth noting that the SVR model was calibrated at these 18 control points and it is expected to perform better. The SVR approach requires the use of 18 separately trained SVR models to make predictions for these locations while a single CNN model (CNN-2005) is sufficient to make comparable predictions, showing clear advantage in real-world applications.

Table 3. The average NSEs and RMSEs calculated by the CNN-2005 and SVR models at the 18 control locations.

| | **CNN-2005** | | **SVR** | |
|---|---|---|---|---|
| Error measure | NSE | RMSE (m) | NSE | RMSE (m) |
| Average value | 0.86 | 0.11 | 0.91 | 0.08 |

### *4.2 Comparing the inundation maps predicted by CNN-2005, SVR and LISFLOOD-FP for the 2005 event*

Fig. 9 presents four sets of predicted inundation maps at a time interval of 12 hours to represent the early, growing, peak and receding stages of the 2005 Carlisle flood event. The flood peak is predicted by all three models to occur at about 12:00 hours on 8$^{th}$ January 2005, which is consistent with the actual event. From the results, it is evident that the inundation maps produced by the CNN-2005 are more effective in capturing the inundated areas compared to the interpolated maps from the SVR models. The CNN-2005 inundation maps are almost identical to those predicted by the LISFLOOD-FP in all four flood stages, suggesting that the model is able to accurately emulate the time-varying water depths for the entire domain. Whilst the SVR approach predicts a satisfactory peak flood map in comparison to the other two models, it evidently overestimates the inundated area for the flood growing and receding periods. In addition, some unexpected small patches of water can be detected in the SVR flood maps throughout the event. This is probably the numerical error caused by the RK interpolation which considers surface elevation of the ground (i.e. bed elevation) as auxiliary information (the correlation between water depth and surface elevation) and may calculate incorrect depth values at those cells with low surface elevation, irrespective of flooding conditions.

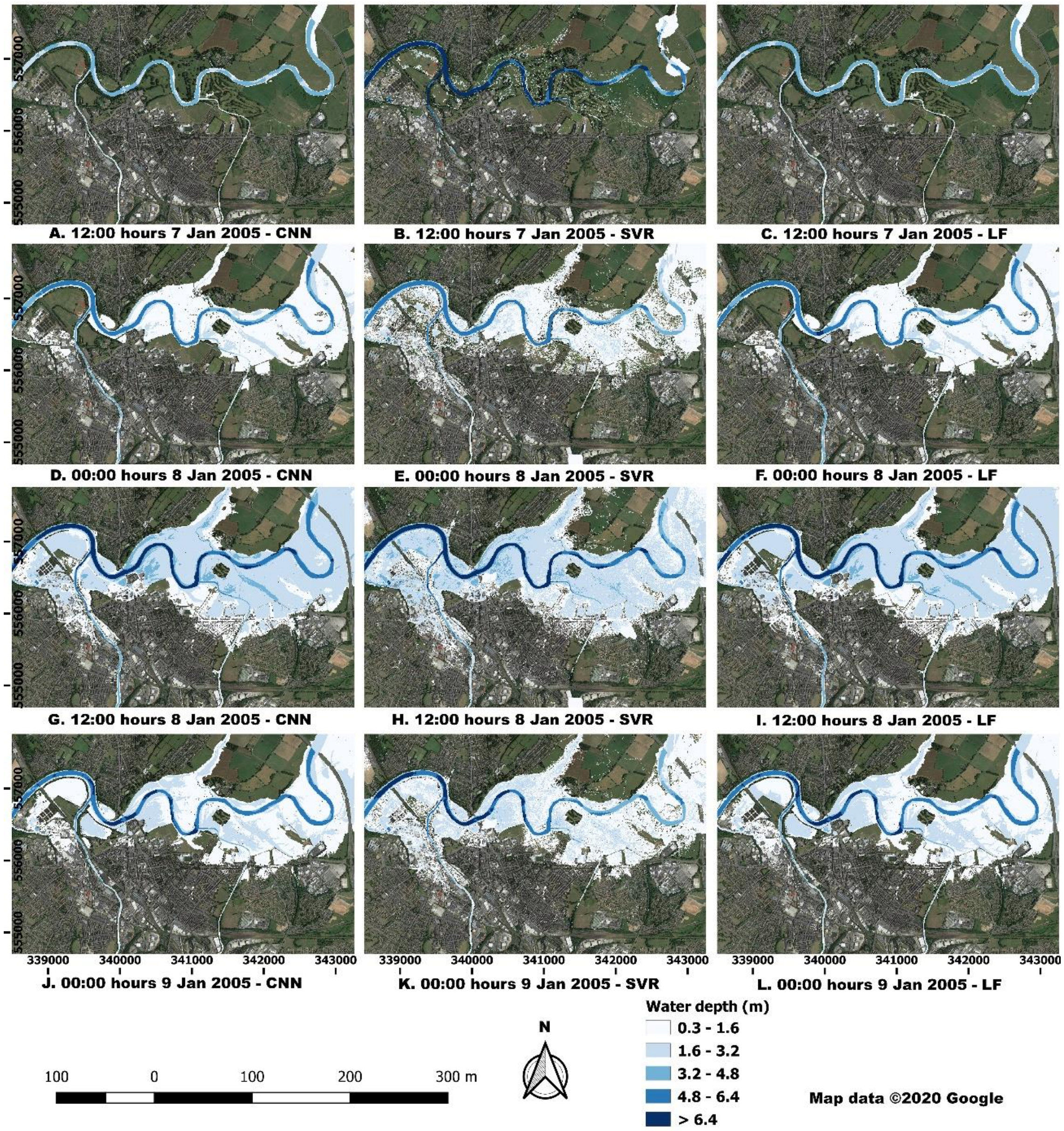

Figure 9. The flood maps created from the CNN-2005, SVR and LISFLOOF-FP predicted water depths during different stages of the flood event.

The descriptive statistics for the inundation maps predicted by the three models are presented in Table 4, showing that both of the CNN and SVR methods produce results that are consistent with the LISFLOOD-FP outputs for all four flood stages. It is interesting to note that although the CNN-2005 predictions visually appear to be much better than the SVR results (Fig. 9), this does not seem to be reflected in the descriptive statistics.

Table 4. Descriptive statistics for the CNN-2005, SVR and LISFLOOD-FP predictions at different stages of the event.

| **Date** | **Model** | **Max (m)** | **Mean (m)** | **St. dev. (m)** |
|---|---|---|---|---|
| 12:00 | CNN-2005 | 4.71 | 0.11 | 0.57 |
| 7th Jan 2005 | SVR | 4.28 | 0.11 | 0.57 |
| | LISFLOOD-FP | 4.95 | 0.12 | 0.62 |
| 00:00 | CNN-2005 | 6.49 | 0.36 | 0.97 |
| 8th Jan 2005 | SVR | 6.41 | 0.38 | 0.97 |
| | LISFLOOD-FP | 6.61 | 0.38 | 1.00 |
| 12:00 | CNN-2005 | 8.17 | 0.85 | 1.47 |
| 8th Jan 2005 | SVR | 8.11 | 0.81 | 1.42 |
| | LISFLOOD-FP | 8.10 | 0.81 | 1.44 |
| 00:00 | CNN-2005 | 7.11 | 0.50 | 1.11 |
| 9th Jan 2005 | SVR | 7.10 | 0.51 | 1.11 |
| | LISFLOOD-FP | 7.19 | 0.53 | 1.14 |

To further investigate the robustness of the models, the spatial error maps (i.e. the absolute difference of water depth between the ML and LISFLOOD-FP inundation maps) for all four stages are produced and presented in Fig. 10, in which the first column shows the error ranges from the CNN-2005 predictions and the second column demonstrates the SVR error maps. From the error maps, it is clear that CNN-2005 produces flood maps matching much more closely with the reference LISFLOOD-FP flood maps, with significantly smaller prediction errors. The resulting maximum error (1.2m) of the CNN-2005 model is almost at a similar level of the lowest error range of the SVR interpolated maps of water depth.

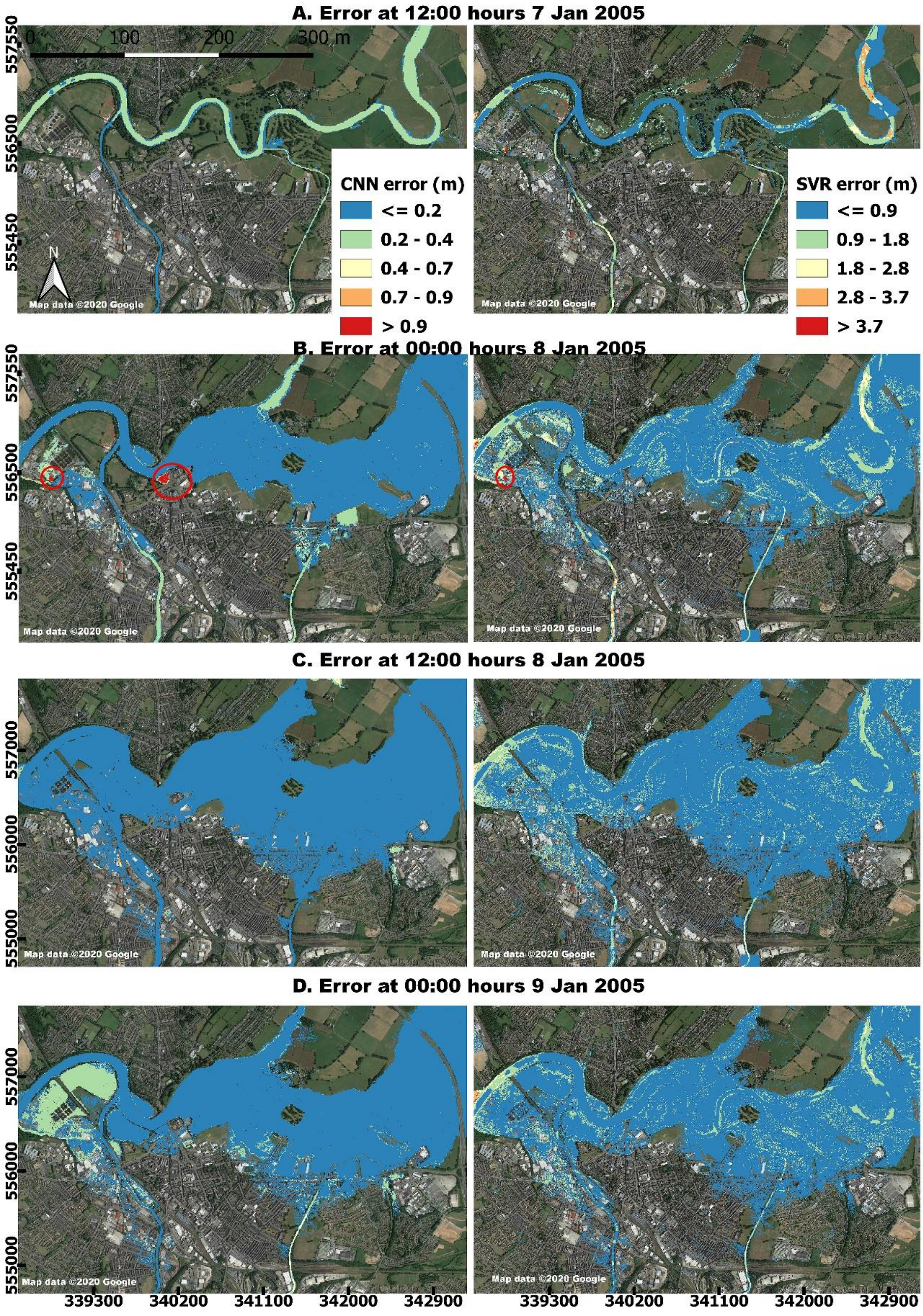

Figure 10. Spatial distribution of the model errors at the four flood stages, column 1: CNN-2005 predictions; column 2: SVR results.

During the peak flood stage (12:00 hours on 8th January), the maximum error of the CNN-2005 prediction is 0.8m and overall maximum prediction error of 1.2m occurs at 00:00 hours on 8th January. These big simulation errors are found to occur at the two specific locations near to Hardwicke Circus (Centre) and Energi Trampoline Park (West) (marked by red circles in Fig. 10). To further investigate the performance of the models after removing these singular points from the statistics, the 99th percentile prediction error during the peak is calculated for CNN-2005, which is 0.2m. This means that 99% of the cells inside the simulation domain have an error less or equal to 0.2m when predicting water depth at the peak flood stage. Similarly, the 99th percentile error of the CNN-2005 for flood growing and receding periods is 0.4m. However, the 99th percentile errors of the SVR model in predicting the water depth at the peak, growing and receding periods are 1.6m, 1.9m and 1.5m, respectively. The small prediction errors of CNN-2005 in all flood stages confirm that the model is more effective in correctly predicting the spatially varying water depths.

To continue on quantifying the performance of the ML predictive models in reproducing the spatial predictions of the LISFLOOD-FP, the flood extents are derived for different stages of the flood event and the number of cells being correctly classified as flooded is counted to calculate the *recall, precision* and *F1* scores, which are summarised in Table 5. Herein, a threshold value of 0.3m is used to delineate wet-dry cells. From the statistic scores, it is clear that almost all of the cells (nearly 100%) classified by the LISFLOOD-FP as flooded are correctly captured by CNN-2005. As a summary, the CNN-2005 emulates the outputs from the LISFLOOD-FP to a high level of accuracy throughout the whole flood event and outperforms the SVR model to a large margin. It may be concluded that, whilst the results predicted by the SVR model may be of higher accuracy at sampled points/cells (Sect. 4.1), the model's accuracy is drastically reduced when interpolation is conducted to produce the spatially distributed depth maps. Whilst it may be possible to enhance the accuracy of the SVR model by increasing the number of samples, this will subsequently increase the number of SVR models that need to be trained and therefore require more computational resources. On the other hand, the CNN model predicts spatially varying water depth at a much higher level of accuracy and a single model is sufficient for city-scale inundation modelling, proving great potential for real-world applications.

Table 5. Spatial accuracy scores of the CNN-2005 and SVR models against LISFLOOD-FP during the flood initiation, growing, peak and recession stages.

| **Date** | **Precision** | | **Recall** | | **F1** | |
|---|---|---|---|---|---|---|
| | **CNN-2005** | **SVR** | **CNN-2005** | **SVR** | **CNN-2005** | **SVR** |
| 12:00 7$^{th}$ Jan 2005 | 0.99 | 0.59 | 0.96 | 0.74 | 0.98 | 0.65 |
| 00:00 8$^{th}$ Jan 2005 | 0.98 | 0.75 | 0.97 | 0.87 | 0.98 | 0.81 |
| 12:00 8$^{th}$ Jan 2005 | 0.98 | 0.90 | 0.99 | 0.97 | 0.99 | 0.93 |
| 00:00 9$^{th}$ Jan 2005 | 0.99 | 0.88 | 0.96 | 0.94 | 0.98 | 0.90 |

Finally, the CNN-2005 predicted peak flood extent at 12:00 hours on January 8$^{th}$ is overlaid with the extents derived from the LISFLOOD-FP outputs and also EA post-event survey in Fig. 11. Overall, the CNN-2005 flood extent matches well with that produced by the LISFLOOD-FP, which is expected and a high classification accuracy between the two sets of model outputs is found (Table 5). However, slight discrepancies can be seen between the CNN-2005 and the surveyed flood extents near to Denton Holme (close to Upstream 3), Melbourne Park (left bank of River Petteril), Willow Holme (confluence of River Caldew and Little Caldew) and Warwick Road East and West. This may be attributed to the assumptions we made in upstream boundary conditions (the use of approximated hydrographs at the boundary points (Upstream 1, 2 and 3) to drive the flood simulations). It should be noted that the CNN model is trained to emulate the LISFLODD-FP results in this work, which is deemed to be highly successful at this sense.

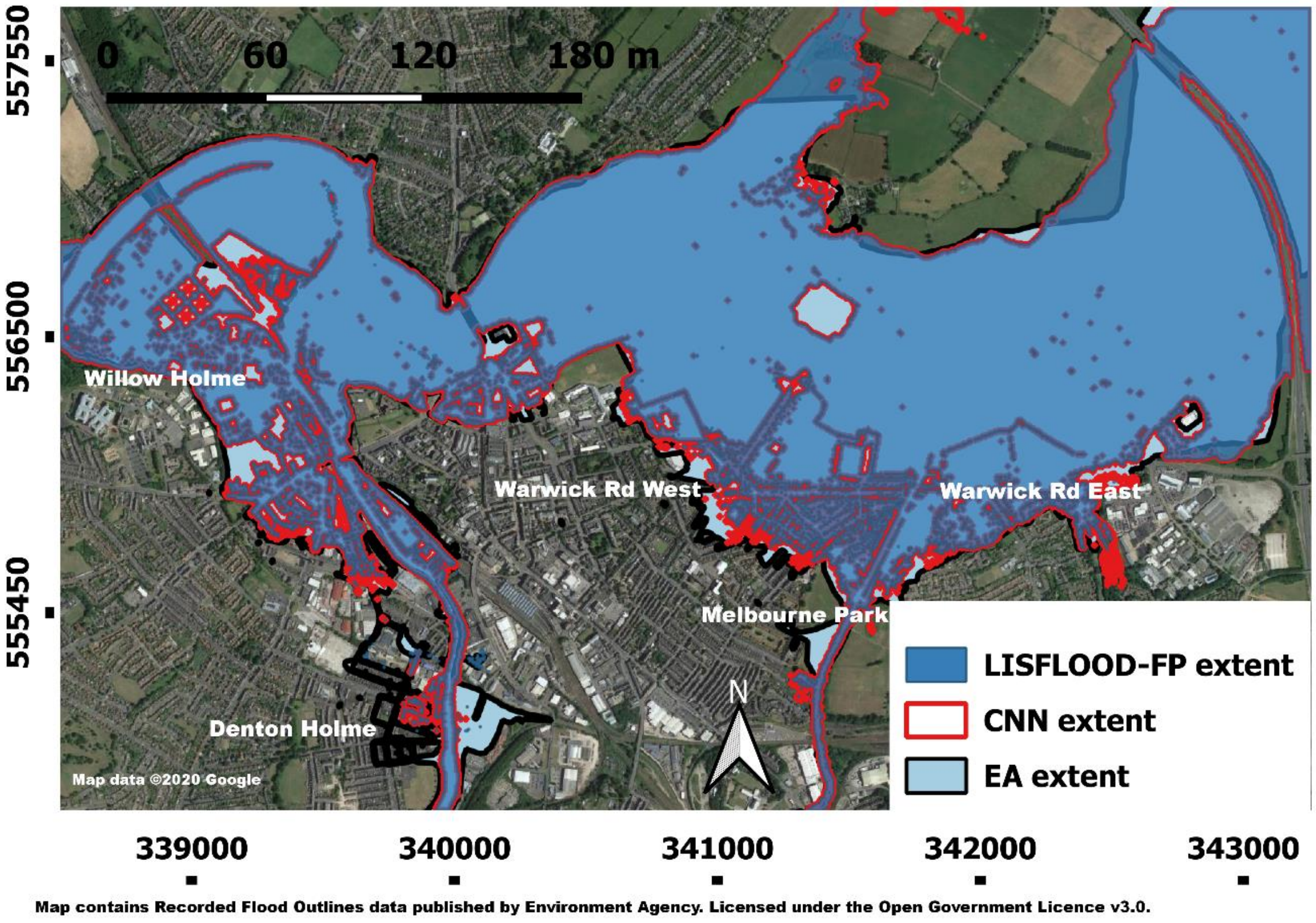

Figure 11. Comparing the CNN-2005 derived flood extent with the LISFLOOD-FP and surveyed extents.

### *4.3 Comparison between CNN-2015 and LISFLOOD-FP for the 2015 event*

It has been demonstrated that the CNN model is able to emulate the LISFLOOD predictions to a high level of accuracy for the 2005 Carlisle flood and the model outperforms the SVR method in predicting spatially distributed water depths. In this section, we further confirm the effectiveness of the CNN model as an emulator of a hydraulic model for more complex flood dynamics that is influenced by flood defences (i.e. changes of domain topography). The magnitude of the 2015 flood event exceeded the design criteria of flood walls and embankments constructed after the 2005 event. The existence of the flood defence system might effectively delay the arrival of the flood and leave more time for the emergency responders to take necessary actions to mitigate the potential impact (Cumbria City Council, 2017). Herein, we conduct numerical experiments to examine the ability of the CNN model (i.e. CNN-2015), after being trained, in capturing the change of flood dynamics as induced by the topographic changes.

Fig. 12 presents the flood maps produced by the trained CNN-2015 model for the early, growing, peak and receding stages of the event. From the results, the impact of the flood defences can be clearly seen during the early and growing periods of the event (i.e. 10:00 hours on 5th December and 18:00 hours on 5th December, respectively). The flood water is held back by the defences at these early stages, leading to a delay in inundation, which has been successfully captured by CNN-2015. Once the flood defences are overtopped, severe inundation occurs and evolves rapidly, which is again reliably predicted by both of the CNN-2015 and LISFLOOD-FP models. Quantitively, the mean NSE and RMSE calculated against the time series of water depth at the control points are respectively 0.94 and 0.18m, confirming CNN-2015 successfully emulates the outputs of the LISFLOOD-FP to a very high level of accuracy.

Fig. 13 shows the error maps of the CNN-2015 predictions against the LISFLOOD-FP outputs for all four flood stages. Similar to the CNN-2005 results, the maximum error (2.2m) occurs during the flood growing stage at the same locations (i.e. near to Hardwicke Circus and Energi Trampoline Park) and the maximum prediction error during the peak (06:00 hours 6th December) is estimated to be 0.8m. The 99th percentile errors calculated for the early, growing, peak and receding periods are 0.5m, 1m, 0.5m and 0.3m, respectively. Table 6 presents the *recall*, *precision* and *F1* scores for the simulation, confirming that CNN-2015 is highly accurate in correctly classifying the wet/dry cells at all stages of the event.

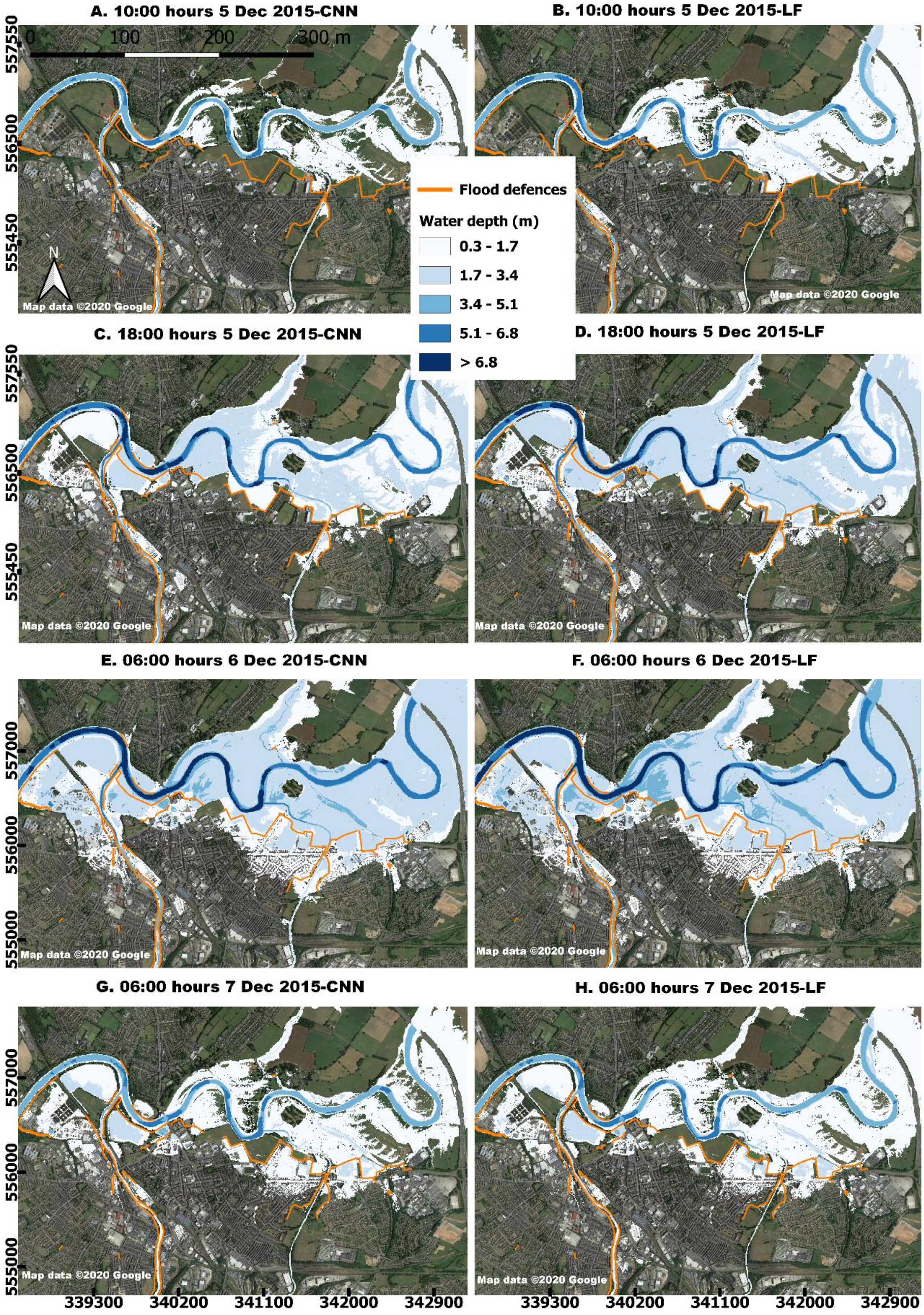

Figure 12. The CNN-2015 and LISFLOOD-FP predicted flood maps, revealing the influence of the flood defences at different stages of the 2015 flood event on 5th and 6th December.

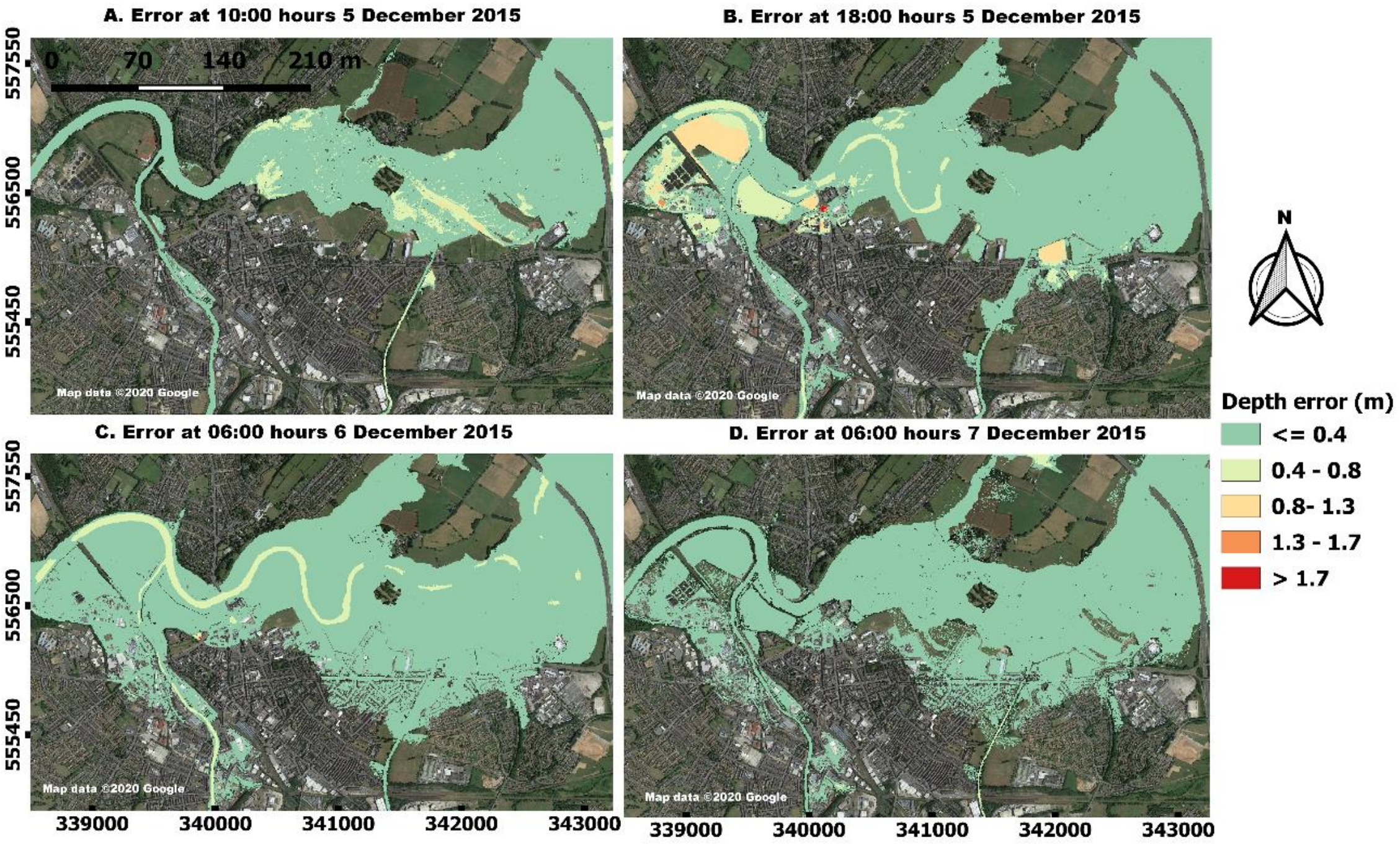

Figure 13. Spatial distribution of the CNN-2015 model errors at the four flood stages.

Table 6. Classification accuracy scores of CNN-2015 against LISFLOOD-FP in predicting the 2015 flood event.

| Date | Precision | Recall | F1 |
|---|---|---|---|
| 10:00 5th Dec 2015 | 0.99 | 0.89 | 0.94 |
| 18:00 5th Dec 2015 | 0.99 | 0.94 | 0.97 |
| 06:00 6th Dec 2015 | 0.99 | 0.99 | 0.99 |
| 06:00 8th Dec 2015 | 0.99 | 0.93 | 0.96 |

To investigate the ability of CNN-2015 in capturing the defence effect on intervening the flood dynamics, the four-stage inundation maps from the simulation without considering the defences are produced and presented in Fig. 14. Comparing with flood maps in Fig. 12 where flood defences are embedded in the simulation, it can be clearly seen that a larger extent of the domain is inundated during the flood growing stage (Fig. 14B) compared with the result in Fig. 12C. Also, an inverse effect of the defences is detected during the flood receding stage, where the water is held back by the defences in the floodplain for a longer duration. At the flood peak, the inundation maps predicted by the two models (with and without considering defences) become more consistent, which is expected.

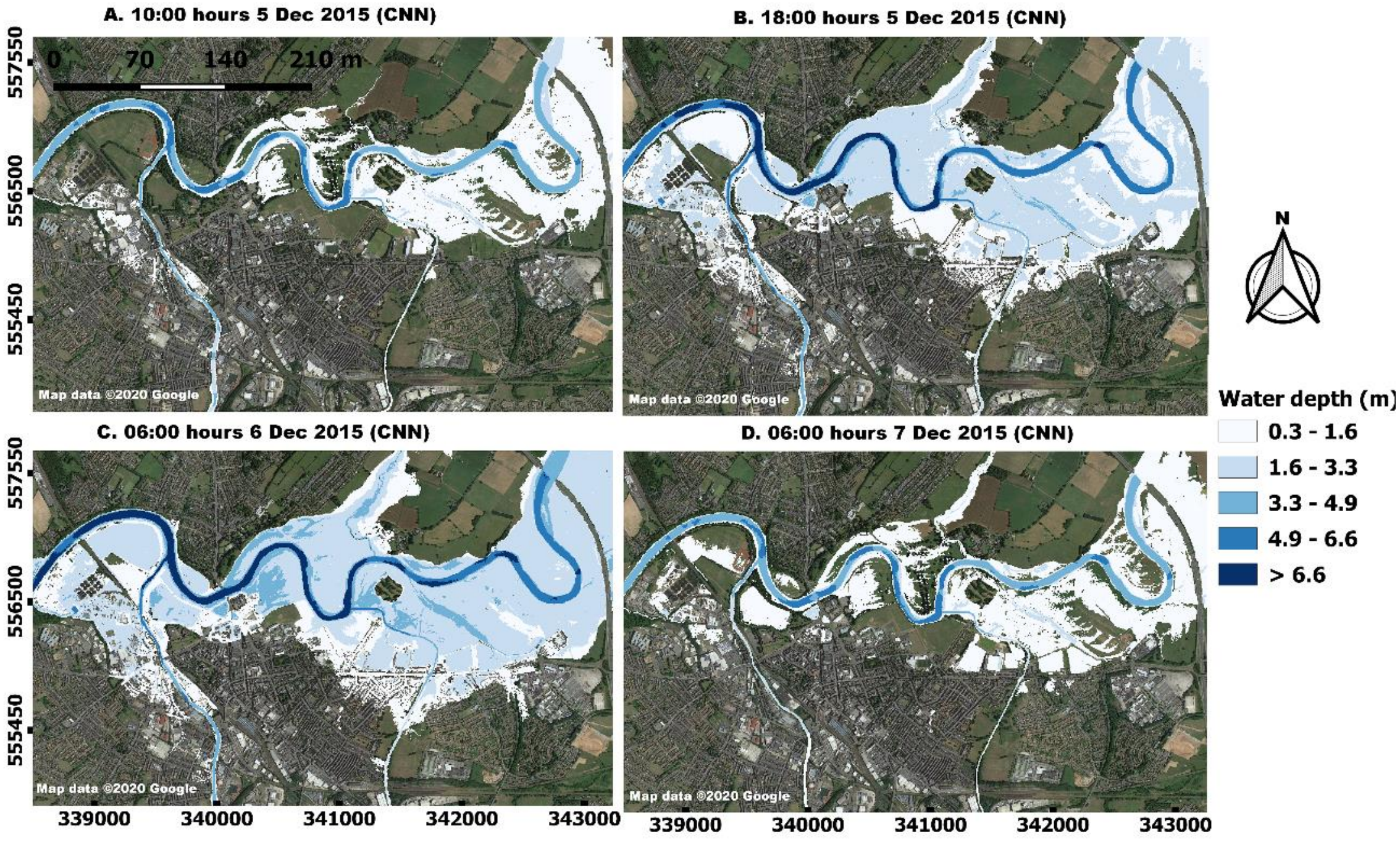

Figure 14. The flood maps predicted by the CNN model for the 2015 event without considering flood defences.

### *4.4 Computational time*

Table 7 provides the run times required by CNN-2005 and LISFLOOD-FP to reproduce the 2005 Carlisle flood. For the CNN model, it takes only a couple of minutes to generate the sequential water depth maps for the entire domain and a few minutes to train the model. It should be noted that, once trained, the CNN model can be used to efficiently predict different flood scenarios. Drawing a completely fair comparison between the two modelling approaches is difficult since they use different computing platforms. Nevertheless, the numbers in Table 7 can still indicate that the CNN model is computationally efficient for real-time applications and can potentially be further explored for supporting probabilistic flood forecasts using ensemble numerical weather predications.

Table 7. The runtimes of the CNN-2005 and LISFLOOD-FP models for simulating the 2005 Carlisle flood event.

| Model | Train time | Total time required to simulate entire event | Output format | Computing device |
|---|---|---|---|---|
| CNN | 3 minutes | 2 minutes | Raster | NVIDIA Tesla P100 GPU |
| LISFLOOD-FP | — | 77 minutes | Raster | Intel I5-9400 2.90GHz CPU |

## 5. Conclusions

The need for evidence-based flood management is greater than ever before due to rapid urbanisation and climate change that have already led to increased flood risk across the world. In this context, fast, reliable and robust modelling tools for real-time flood prediction/forecasting is important for assessing the multidimensional social and economic impacts of and providing reliable forecasts to enhance societal resilience to flooding. This work introduces a deep CNN approach for rapid fluvial flood modelling that can potentially be leveraged for operational flood nowcasting or forecasting. The idea underpinning the proposed study is that cell-based water depths in a floodplain are a function of time varying discharge and the time of observation at the upstream. Therefore, a non-linear function can be fitted between sequences of historical geographically distributed water levels/depths and observed discharges to predict water levels for the future flood events. The results show that the CNN model can effectively emulate the outputs (i.e. water depths) of a 2D hydraulic model with a high level of accuracy. It also shows that the model can be easily re-trained to account for the major topographical change (e.g. construction of flood defences) and capture the resulting influence on flood dynamics. Compared with traditional ML methods such as SVR, a single CNN model has the capability to make predictions for a domain consisting hundreds of thousands of cells. Due to the high computational efficiency, superior performance and simplicity, the proposed method offers a promising tool for real-time nowcasting/forecasting of flood inundation.

## Model and data availability

The models constructed in this paper and reproduceable data have been made available through https://github.com/SRKabir/Rapid_FloodModelling_CNN.

## Acknowledgements

This work is partly funded by the Newton Fund and UK Met Office 'WCSSP-India Lot 7: Building a Flood Hazard Impact Model for India (FHIM-India) (DN394978)' project.